# scientific reports

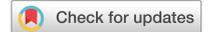

**OPEN**   A novel framework for spatio-temporal prediction of environmental data using deep learning

Federico Amato[1]✉, Fabian Guignard[1], Sylvain Robert[2] & Mikhail Kanevski[1]

As the role played by statistical and computational sciences in climate and environmental modelling and prediction becomes more important, Machine Learning researchers are becoming more aware of the relevance of their work to help tackle the climate crisis. Indeed, being universal nonlinear function approximation tools, Machine Learning algorithms are efficient in analysing and modelling spatially and temporally variable environmental data. While Deep Learning models have proved to be able to capture spatial, temporal, and spatio-temporal dependencies through their automatic feature representation learning, the problem of the interpolation of continuous spatio-temporal fields measured on a set of irregular points in space is still under-investigated. To fill this gap, we introduce here a framework for spatio-temporal prediction of climate and environmental data using deep learning. Specifically, we show how spatio-temporal processes can be decomposed in terms of a sum of products of temporally referenced basis functions, and of stochastic spatial coefficients which can be spatially modelled and mapped on a regular grid, allowing the reconstruction of the complete spatio-temporal signal. Applications on two case studies based on simulated and real-world data will show the effectiveness of the proposed framework in modelling coherent spatio-temporal fields.

Data science plays a primary role in tackling climate and environmental challenges[1,2]. While the amount of observations from earth-observing satellites and in-situ weather monitoring stations keeps growing, climate modelling projects are generating huge quantities of simulated data as well. Analysing these data with physically-based models can be an extremely difficult task. Consequently, climate change researchers have become particularly interested in the role played by Machine Learning (ML) towards the advances of the state-of-the-art in climate modelling and prediction[3]. In parallel, ML researchers are becoming aware of the relevance of their work to help tackle the environmental and climate crisis[4].

Environmental spatio-temporal data are usually characterized by spatial, temporal, and spatio-temporal correlations. Capturing these dependencies is an extremely important task. Deep Learning (DL) is a promising approach to tackle this challenge, especially because of its capability to automatically extract features both in the spatial domain (with Convolution Neural Networks (CNNs)) and in the temporal domain (with the recurrent structure of Recurrent Neural Networks (RNNs)). However, significant efforts must still be spent to adapt traditional DL techniques to solve environmental and climate-related spatio-temporal problems[5].

Recent years have seen numerous attempts to use DL to solve environmental problems, such as precipitation modelling[6,7] or extreme weather events[8] and wind speed forecasting[9]. However, until now most of the efforts have been focused on transposing methodologies from computer vision to the study of climate or environmental raster data—i.e. measurements of continuous or discrete spatio-temporal fields at regularly fixed locations[10]. These data are generally coming from satellites for earth observation or from climate models outputs. Nevertheless, especially at the local or regional scales, there could be several reasons to prefer to work with direct ground measurements—i.e. measurements of continuous spatio-temporal fields on a set of irregular points in space—rather than with raster data[11]. Indeed, ground data are more precise than both satellite products and climate models output, generally with a higher temporal sampling frequency of phenomena and without missing data due to e.g. clouds. On the other hand, the direct adaptation of models coming from traditional DL is more complex in the case of spatio-temporal data collected over sparsely distributed measurement stations. In this case, the approaches discussed in literature to date only permit to perform modelling at the locations of the measurement stations and, up to our knowledge, no ML-based methodology has been proposed to solve spatio-temporal interpolation problems having as input spatially irregular ground measurements.

[1]Faculty of Geosciences and Environment - Institute of Earth Surface Dynamics, University of Lausanne, Lausanne, Switzerland. [2]Swiss Re, Zurich, Switzerland. ✉email: federico.amato@unil.ch

To tackle this gap, we propose an advanced methodological framework to reconstruct a spatio-temporal field on a regular grid using spatially irregularly distributed time series data. The spatio-temporal process of interest is described in terms of a sum of products between temporally referenced basis functions and corresponding spatially distributed coefficients. The latter are considered as stochastic and the problem of the estimation of the spatial coefficients is reformulated in terms of a set of regression problems based on spatial covariates, which are then learned jointly using a multiple output deep feedforward neural network. The proposed methodology permits to model non-stationary spatio-temporal processes.

Across two different experimental settings using both simulated and real-world data, we show that the proposed framework allows to reconstruct coherent spatio-temporal fields. The results have practical implications, as the methodology can be used to interpolate ground measurements of climate and environmental variables keeping into account the spatio-temporal dependencies present in the data.

The remainder of this paper is structured as follows. First, the literature related to our study is reviewed. Subsequently, the framework for the decomposition of spatio-temporal data and for the modelling of the obtained random spatial coefficients is introduced, elaborating on the technical details with respect to the regression models used in the experiments. The proposed methodology is then tested on synthetic and real-world case studies. Finally, results are discussed and future research directions are drawn.

## Related works

In this section, we contextualize our methodological framework in relation to existing work in the field. Classical methods for spatio-temporal modelling include state-space models[12] and Gaussian Processes based on spatio-temporal kernels[13], including their applications in geostatistics[14]. The latter includes the widely adopted kriging models, which can encounter issues in correctly reproducing non-linear spatio-temporal behaviours, as in the case of complex climatic fields. Moreover, the use of kriging models assumes a deep knowledge of the dependence structure present in the data, to be applied to correctly specify its representation via the covariance function used while modelling[15]. Finally, in kriging the use of high number of covariates, despite being possible in what is generally referred to as co-kriging, is often complicated and has some very restrictive constraints. Traditional ML algorithms have also been used to solve spatio-temporal problems[16]. However, these techniques typically require human-engineered spatio-temporal features, while DL can automatically learn feature representations from the raw spatio-temporal data[17].

The motivation for this study originates from recently proposed DL approaches to analyse spatio-temporal patterns. Geospatial problems have been studied with non-Euclideian spatial graphs via graph-CNNs[18]. Generative adversarial networks (GAN) have also been used together with local autocorrelation measures to improve the representation of spatial patterns[19]. Image inpainting techniques have been adapted to the imputation of missing values in gridded spatial climate datasets by reconstructing missing values using stacked partial convolutions and an automatic mask updating mechanism via transfer learning[20]. Still, this approaches did not consider the temporal dimension of the studied phenomena.

When the temporal dimension is considered and the data are collected as raster, the underlying spatio-temporal field can be modelled with techniques which already proved their effectiveness in extracting features from images and videos, like CNNs[21], RNNs[22] or with mixed approaches like in Convolutional Long Short-Term Memory networks[7]. Bayesian methods have also been proposed together with RNNs to quantify prediction uncertainties[23].

Differently, few approaches have been proposed to model data collected at spatially irregular locations. Several studies investigated the ability of ML techniques to solve the problem of imputation of missing data in environmental time series[24,25]. Concerning forecasting problems, a common approach to take into account the correlation among the different measurement locations is to consider them as nodes in a graph, which can then be modelled using specific DL architectures[26–28]. The main limitation of such methodology is that prediction is only possible at the spatial locations of the measurement stations and not at any spatial location of potential interest.

Nonetheless, to the best of our knowledge, no study has been conducted on the possibility of performing interpolation of spatio-temporal data at any spatial location using DL, highlighting the novelty of the framework proposed in this paper.

## Methods

As already discussed, when working with spatio-temporal phenomena it is difficult to realistically reproduce the spatial, temporal and spatio-temporal dependencies in the data. One way of keeping into account these dependencies is to adopt a basis function representation[29]. Numerous basis functions can be used to decompose spatio-temporal data, including complete global functions as with Fourier analysis, and non-orthogonal bases as with wavelets. Here we use the reduced-rank basis obtained through a principal component analysis (PCA), also known as an Empirical Orthogonal Functions (EOFs) decomposition in the fields of climatology, meteorology and oceanography[30].

While in ML PCA is generally applied to reduce the dimensionality of the input space, here we use it on the output space—i.e. the spatio-temporal target variable we want to interpolate—in order to decompose the data into fixed temporal bases and their corresponding spatial coefficients. The latter are then considered as the target variable in a spatial regression problem, which can, in principle, be solved using any ML technique. More specifically, the framework that we propose for the interpolation of continuous spatio-temporal fields starting from measurements on a set of irregular points in space consist of the following steps. First, a basis function representation is used to extract fixed temporal bases from the spatio-temporal observations. Then, the stochastic spatial coefficients corresponding to each basis functions are modelled jointly at any desired spatial location with a DL regression technique. Finally, the spatio-temporal signal is recomposed, returning a spatio-temporal interpolation of the field.

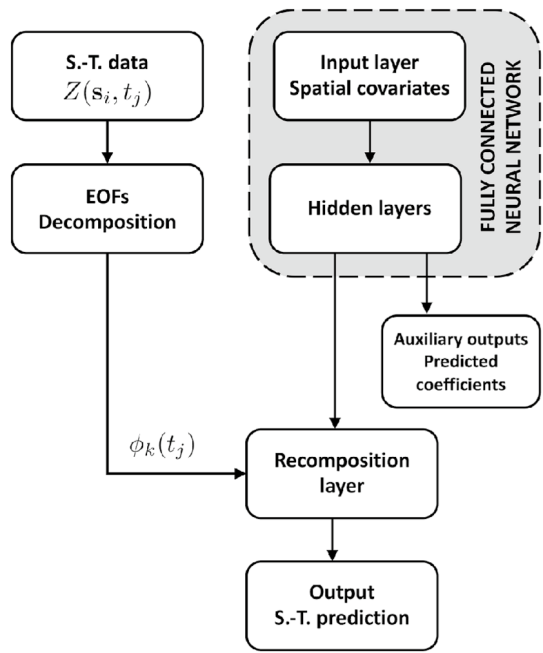

**Figure 1.** The architecture of the proposed framework. The temporal bases are extracted from a decomposition of the spatio-temporal signal using EOFs. Then, a fully connected neural network is used to learn the corresponding spatial coefficients. The full spatio-temporal field is recomposed following Eq. (1) and used for loss minimization.

In the following, we discuss details about the decomposition of the spatio-temporal signal using EOFs and about the structure of the regression problem to be solved to perform the spatial interpolation of the coefficients. We also provide the mathematical formulation of spatio-temporal semivariograms, which will be later used to empirically evaluate the quality of the prediction models.

**Decomposition of spatio-temporal data using EOFs.** Let us suppose that we have spatio-temporal observations $\{Z(\mathbf{s}_i, t_j)\}$ at $S$ spatial locations $\{\mathbf{s}_i : 1 \leq i \leq S\}$ and $T$ time-indices $\{t_j : 1 \leq j \leq T\}$. Let $\tilde{Z}(\mathbf{s}_i, t_j)$ be the spatially centered data,

$$\tilde{Z}(\mathbf{s}_i, t_j) := Z(\mathbf{s}_i, t_j) - \bar{\mu}(t_j),$$

where

$$\bar{\mu}(t_j) := \frac{1}{S} \sum_{i=1}^{S} Z(\mathbf{s}_i, t_j),$$

is the global mean at time $t_j$.

The centred data $\tilde{Z}(\mathbf{s}_i, t_j)$ can be represented with a discrete temporal orthonormal basis $\{\phi_k(t_j)\}_{k=1}^{K}$, i.e.

$$\tilde{Z}(\mathbf{s}_i, t_j) = \sum_{k=1}^{K} \alpha_k(\mathbf{s}_i) \phi_k(t_j), \qquad (1)$$

such that

$$E[\alpha_k(\mathbf{s}_i)] = 0, \text{ for } k = 1, \ldots K,$$
$$\text{Var}[\alpha_1(\mathbf{s}_i)] \geq \text{Var}[\alpha_2(\mathbf{s}_i)] \geq \cdots \geq \text{Var}[\alpha_K(\mathbf{s}_i)] \geq 0,$$
$$\text{Cov}[\alpha_{k_1}(\mathbf{s}_i), \alpha_{k_2}(\mathbf{s}_i)] = 0, \text{ for all } k_1 \neq k_2,$$

where $\alpha_k(\mathbf{s}_i)$ is the coefficient with respect to the $k$-th basis function $\phi_k$ at spatial location $\mathbf{s}_i$, and $K = \min\{T, S-1\}$. Notice that the scalar coefficient $\alpha_k(\mathbf{s}_i)$ only depends on the location and not on time, while the temporal basis function $\phi_k(t_j)$ is independent of space. The decomposition (1) is theoretically justified by the Karhunen-Loève expansion[31], which is based on Mercer's theorem[32].

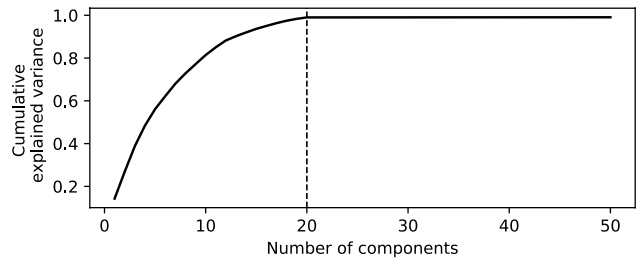

**Figure 2.** Cumulative percentage of variance explained by the first 50 components of the EOFs decomposition for the simulated dataset. As expected from the data generating process, the sum of the relative variances of the 20 first components reach the total variance of the data.

The basis computation is related to the spectral decomposition of empirical temporal covariance matrix. However, a singular value decomposition (SVD) is more efficient in practice. Furthermore, the SVD performed on $(S-1)^{-\frac{1}{2}} \cdot \tilde{Z}(\mathbf{s}_i, t_j)$ yields directly the normalized spatial coefficients[29].

Practically, EOFs computation is equivalent to a PCA[33–35], where time-indices are considered as variables and the realizations of those variables correspond to the spatial realizations of the phenomenon. As in classical PCA, the relative variance of each basis is given by the square of the SVD singular values. Therefore, in case of a reconstruction with a truncated number $\tilde{K} \leq K$ of components, the decomposition (1) compresses the spatio-temporal data and reduces its noise.

**Modeling of the coefficients.** The—potentially truncated—EOFs decomposition returns for each spatial location $\mathbf{s}_i$, corresponding to the original observations, $\tilde{K}$ random coefficients $\alpha_k(\mathbf{s}_i)$. These coefficients can be spatially modelled and mapped on a regular grid solving an interpolation/regression task.

To demonstrate the effectiveness of the proposed approach, we will model the coefficients using a deep feed-forward fully connected neural network[36]. The structure of the network is described in Fig. 1. Spatial covariates are used as inputs for the neural network having a first auxiliary output layer where the spatial coefficients are modelled. A recomposition layer will then use the $\tilde{K}$ modelled coefficients and the temporal bases $\phi_k$ resulting from the EOFs decomposition in order to reconstruct the final output—i.e. the spatio-temporal field—following Eq. (1).

The described network has multiple inputs, namely the spatial covariates—which flow through the full stack of layers—and the temporal bases directly connected to the output layer. It also has multiple outputs, namely the spatial coefficients for each basis, all modelled jointly, and the output signal. While the network is trained by minimizing the loss function on the final output, having as auxiliary output the spatial coefficients maps ensures a better explainability of the model, which is of primary importance in earth and climate sciences.

It has to be highlighted that instead of the proposed structure, any other traditional ML algorithm could have been used to model the spatial coefficients as a standard regression problem, indicating a rather interesting flexibility of the proposed framework. However, the proposed DL approach has two main advantages. The first is that most classical ML regression algorithms cannot handle multiple outputs, and thus one would have to fit separate models for each coefficient map without being able to take advantage of the similarities between the tasks. The second advantage is that our proposed DL approach minimizes the loss directly on the final prediction target, i.e. the reconstructed spatio-temporal field of interest. Differently, when using single output models, each of these is separately trained to minimize a loss computed on the spatial coefficients. This should, in principle, lead to a lower performance than with the proposed DL approach minimizing directly the error computed on the full reconstructed spatio-temporal signal.

**Spatio-temporal semivariograms.** Semivariograms can be used to describe the spatio-temporal correlation structures of a dataset. The (isotropic) empirical semivariogram is given by[29,37]

$$\gamma(h, \tau) = \frac{1}{2 \cdot \#N_s(h) \cdot \#N_t(\tau)} \sum_{\mathbf{s}_i, \mathbf{s}_k \in N_s(h)} \sum_{t_j, t_l \in N_t(h)} \left[ Z(\mathbf{s}_i, t_j) - Z(\mathbf{s}_k, t_l) \right]^2,$$

where $N_s(h)$ is the set of all location pairs separated by a Euclidean distance of $h$ within some tolerance, $N_t(\tau)$ is the set of all time points separated by a temporal lag of $\tau$ within some tolerance and # denotes the cardinality of a set.

In this study, variography is used to understand the quality and the quantity of spatio-temporal dependences extracted by the model from the original data through a residuals analysis[38]. The semivariograms are numerically computed with the gstat R library[39,40].

## Results

The proposed methodology has been applied on two different datasets: a simulated spatio-temporal field, and a real-world dataset of temperature measurements. For both applications we will first introduce the dataset. Data generation/collection as well as the approaches used to allocate training, testing and validation sets are described. Then, the main results from the spatio-temporal prediction approach are presented and discussed.

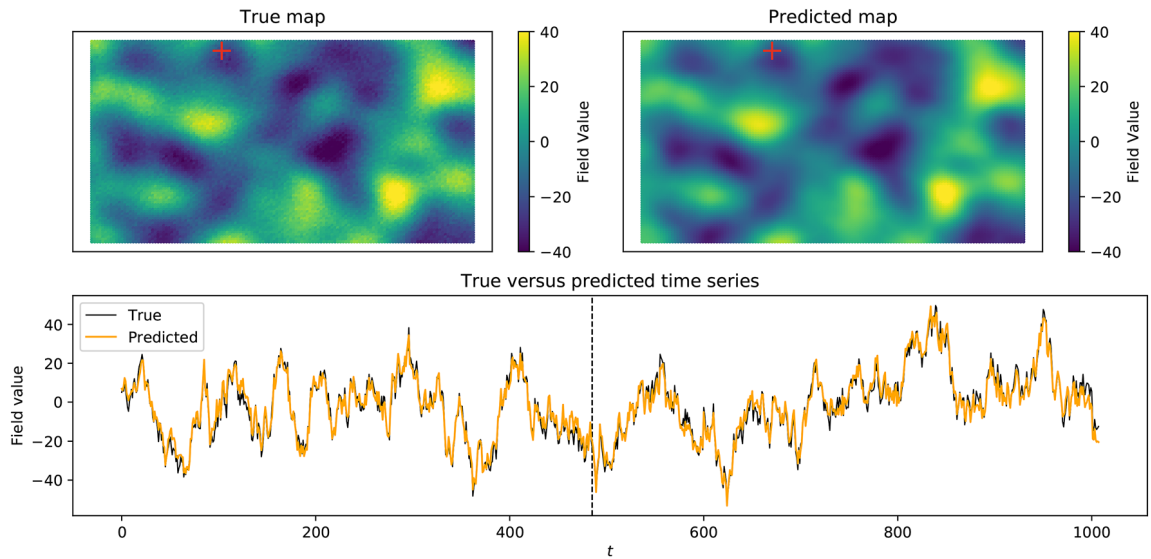

**Figure 3.** Model outputs for the simulated spatio-temporal field defined by Eq. (2). Top left : A snapshot of the true spatial field at the fixed time indicated by the vertical dashed in the temporal plot below. Top right : The predicted map at the same time. Bottom : The true time series (in black) and the predicted time series (in orange) at the fixed location marked by a cross in the maps above.

The spatial coefficient maps are modelled using a fully connected feedforward neural network. The network has been implemented in Tensorflow[41]. For both experiments the following configuration has been adopted. Kernel initializer: He initialization; Activation function: ELU; Regularization: Early Stopping. Optimizer: Nadam; Learning Rate scheduler: 1Cycle scheduling; Number of Hidden Layers: 6; Neurons per layer: 100; Loss function: Mean Absolute Error. Batch Normalization has been used in the real world case study.

**Experiment on simulated dataset.** To produce realistic 2-dimensional spatial patterns, 20 Gaussian random fields with Gaussian kernel are simulated from the RandomFields R package[42] and noted $X_k(\mathbf{s}), k = 1, \ldots, 20$. Time series $Y_k(t_j)$ of length $T = 1080$, for $k = 1, \ldots, 20$, are generated using an order 1 autoregressive model. Then, the simulated spatio-temporal dataset is obtained as a linear combination of the spatial random fields $X_k(\mathbf{s})$, where $Y_k(t_j)$ plays the role of the coefficients at time $t_j$, i.e.

$$Z(\mathbf{s}, t_j) = \sum_{k=1}^{20} X_k(\mathbf{s}) Y_k(t_j) + \varepsilon, \qquad (2)$$

where $\varepsilon$ is a noise term generated from a Gaussian distribution having zero mean and standard deviation equal to the 10% of the standard deviation of the noise-free field. Spatial points $\mathbf{s}_i, i = 1, \ldots, S = 2000$ are sampled uniformly on a regular 2-dimensional spatial grid of size $139 \times 88$, which will constitute the training locations. The spatio-temporal training set $\{Z(\mathbf{s}_i, t_j)\}$ is generated by evaluating the sequence of fields (2) at the training locations. Spatio-temporal validation and testing sets are generated analogously from 1000 randomly selected locations each.

The training and validation sets are decomposed following the methodology described in the previous section. The cumulative percentage of the relative variance for the first 50 components is represented in Fig. 2. Note that without the addition of the $\varepsilon$ term the total variance would have been explained with the first 20 components, which is the number of elements used to construct the simulated dataset, see Eq. (2). Nonetheless, even in the presence of the additional noise these components explain about 99% of the variance. Therefore, the neural network is trained using two inputs, namely the spatial covariates corresponding to the $x$ and $y$ coordinates, and $\tilde{K} = 20$ outputs.

The Mean Absolute Error (MAE) computed on the test set after the modeling with the proposed neural network approach is indicated in Table 1. We also compare the output of the proposed framework to the results obtained using a different strategy. Specifically, instead of predicting all the spatial coefficients at once with the proposed multiple outputs model, we investigated the impact in terms of test error performance due to the separate modeling of each coefficient map. To this aim, both fully connected feedforward neural network (NN) and Random Forest (RF)—which is popular in environmental and climatic literature—have been used to predict the individual spatial coefficient maps, which are then used together with the temporal bases to reconstruct the spatio-temporal field. The neural network has the same structure as the one used with the multiple output strategy, with the exception of the recomposition layer, which is indeed absent. The RF models were implemented in Scikit-learn[43] and trained using 5-fold cross validation. It is shown that the use of the proposed multiple output model helps to significantly improve the performances with respect to the approaches based on separate single output models, which have significantly worse performances.

| Dataset | Multiple outputs | Single output (NN) | Single output (RF) |
|---|---|---|---|
| Simulated | 1.978 | 8.340 | 6.709 |
| Temperature (all comp.) | 1.148 | 1.599 | 1.682 |
| Temperature (24 comp.) | 1.285 | 1.628 | 1.683 |

**Table 1.** Comparison of the test MAE resulting from a model following our novel framework—in which all the spatial coefficients are learned at once with a multiple output deep neural network—and the one obtained through an approach in which each coefficient map is predicted using a separate single output regression model.

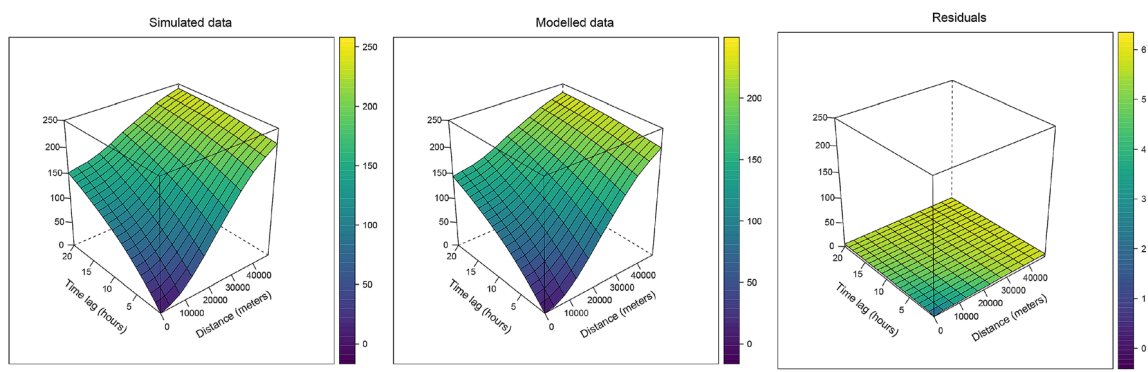

**Figure 4.** Variography for the simulated dataset model. Spatio-temporal semivariograms have been computed on the test points for the simulated data (left), the model implemented using the first 20 EOFs components (center) and its residuals (right).

Randomly chosen examples of prediction map and time series are shown and compared with the true spatio-temporal field in Fig. 3. The predicted map recovers the true spatial pattern, and the temporal behaviours are fairly well replicated too.

Figure 4 shows the spatio-temporal semivariograms of the simulated data, of the output of the model and of the residuals—i.e. the difference between the simulated data and the modelled one. All semivariograms are computed on the test points. The semivariogram on the modelled data shows how the interpolation recovered the same spatio-temporal structure of the (true) simulated data, although its values are slightly lower. This imply that the model has been able to explain most of the spatio-temporal variability of the phenomenon. However, it must be pointed out that even better reconstruction of the spatio-temporal structure of the data could be recognizable in the semivariograms computed on the training set, similarly to how the training error would be lower than the testing one. Finally, almost no structure is shown in the semivariogram of the residuals, suggesting that almost all the spatially and temporally structured information—or at least the one described by a two-point statistic such as the semivariogram—has been extracted from the data. It also shows a nugget corresponding to the noise used in the generation of the dataset.

**Experiment on temperature monitoring network.** The effectiveness of the proposed framework in modelling real-world climate and environmental phenomena is tested on a case study of air temperature prediction in a complex Alpine region of Europe. This area (having projected bounds 486660, 77183, 831660, 294683 meters in the spatial reference system CH1903/LV03) constitutes a challenging—but representative—example. Indeed, the region is traversed from south-west to north-east by the Alps chain. This forms a natural barrier, which leads to marked differences in temperature between the two sides of the mountain range. Starting from measurements sampled with hourly frequency from $1^{st}$ July 2016 to $30^{th}$ June 2018 over 369 meteorological stations, whose spatial distribution represents different local climates of the complex topography of the region, we will model the spatio-temporal temperature field on a regular grid with a resolution of 2500 meters. Data were downloaded from MeteoSwiss (https://gate.meteoswiss.ch) and include measurements from several meteorological monitoring networks of Switzerland, Germany and Italy. The data were thoroughly explored and obvious outliers were removed. Missing data, corresponding to approximately 1% of the entire dataset, were replaced by a local spatio-temporal mean obtained from the values of the eight stations closest in space over two contiguous timestamps, yielding an average over 24 spatio-temporal neighbours[44,45]. Before modelling, the dataset was randomly divided into training, validation and testing subsets, consisting of 220, 75 and 74 stations respectively.

The first three components resulting from the EOFs decomposition of the training and validation sets, together with the corresponding temporal bases functions and normalized spatial coefficients, are displayed in Fig. 5. The first two temporal bases clearly show yearly cycles, and a closer exploration of the time series would reveal other structured features such as daily cycles—here not appreciable because of the visualisation of a large amount of time-indices. The normalized spatial coefficient maps unveil varying patterns at different spatial scales.

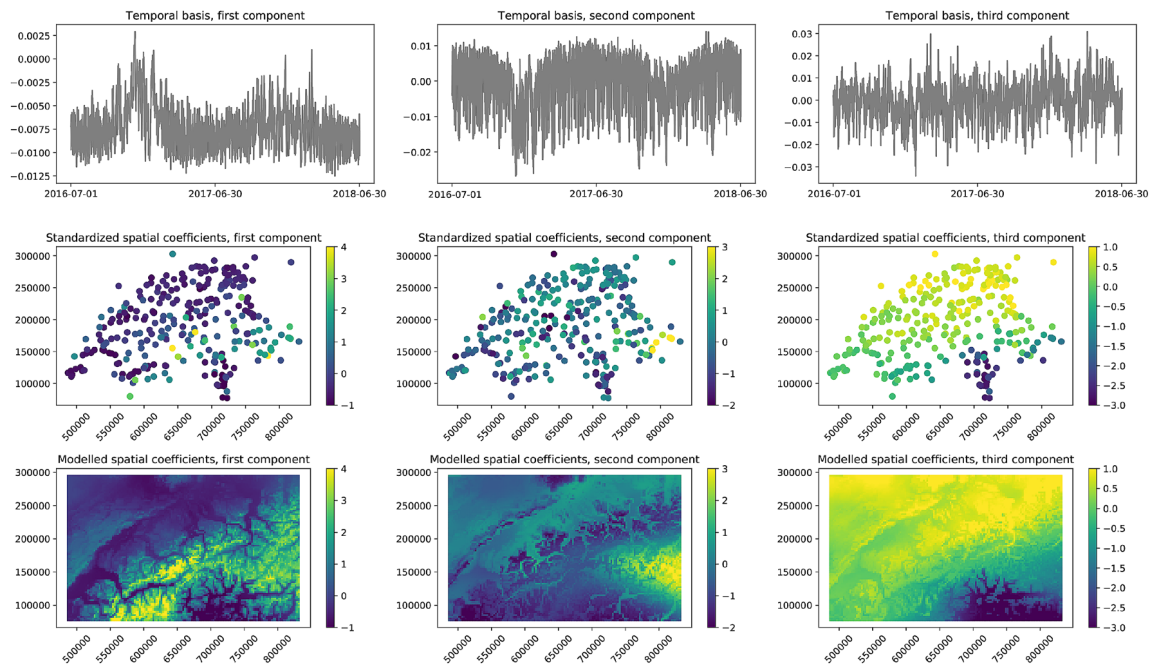

**Figure 5.** Temperature monitoring network, first 3 components of the EOFs decomposition. Top row : The temporally referenced basis functions. Center row : The normalized spatial coefficients of the corresponding EOFs. Bottom row: The corresponding predicted spatial coefficients provided by the auxiliary outputs of the fully connected neural network (all components).

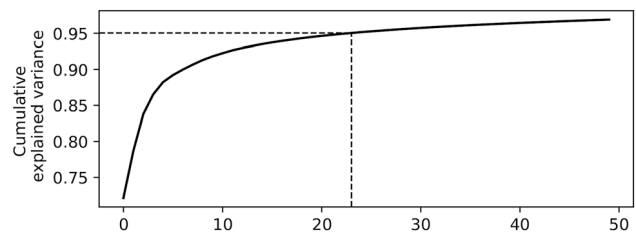

**Figure 6.** Cumulative percentage of variance explained by the first 50 components of the EOFs decomposition for the temperature monitoring network dataset. The sum of the relative variance of the first 24 components reaches 95% of the total data variation.

Fig. 6 shows that 95% of the data variation is explained by the first 24 components. Hence, we implemented two different models. The first one is developed by using all the available components ($\tilde{K} = K = 294$), while the second one adopts a compressed signal keeping 95 % of data variance ($\tilde{K} = 24$). It is well known that temperature strongly depends on topography, and in particular on elevation[46]. Hence, in addition to latitude and longitude, altitude is added as a spatial covariate in the two models. The trained models will therefore have three inputs, i.e. the three spatial covariates, and $\tilde{K}$ outputs, with the latter changing in the two experiments conducted.

Predicted maps of temperature at a randomly chosen fixed time are shown in Fig. 7 for both model, together with time series at a random testing station. The predicted temperatures at the testing station are compared with the true measurements on accuracy plots and on a time series plot. The model with $\tilde{K} = 294$ replicates extremely well the temperature behaviour. The predicted map captures the different climatic zones, while the predicted time series retrieves very well the temporal dependencies in the data. In particular, highly structured patterns such as daily cycles are recovered as well as the variability at smaller temporal scales and abrupt behaviour changes. The accuracy plot further highlights how well the predictions fit the true values. The model with $\tilde{K} = 24$ shows comparable results while the dimensionality of the data has been significantly reduced, indicating that it is possible to obtain similar accuracy with compressed data.

Even if not strictly required to model the spatio-temporal field, the spatial coefficient maps can be obtained from the neural network as auxiliary outputs (shown in Fig. 5). Their usage is extremely relevant from a diagnostic and interpretation perspective. Indeed, climate and Earth system scientists are accustomed to the use of EOFs as exploratory data analysis tool to understand the spatio-temporal patterns of atmospheric and environmental phenomena. Hence, the full reconstruction of the spatial coefficients on a regular grid represents a useful step towards a better explainability of the modelled phenomena. Specifically, these maps could be interpreted from a physical standpoint to analyze the contribution of each temporal variability pattern[29,31]. In our case study, the globally emerging structures correspond to different known climatic zones. As an example, the first map in the bottom row of Fig. 5 clearly shows the contribution of the first temporal basis in the Alps chain, while the third map indicates a strong negative contribution of the corresponding temporal basis at the south of the chain.

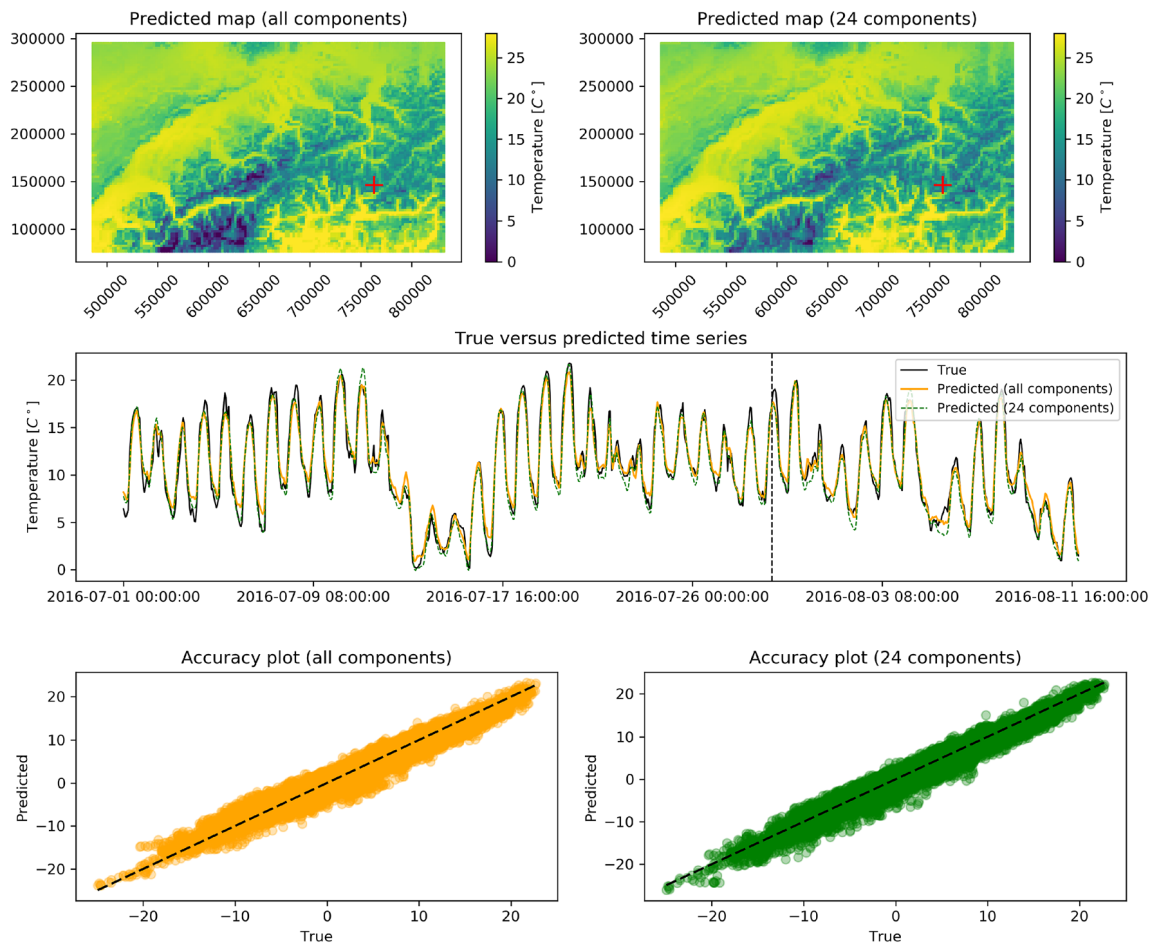

**Figure 7.** Model outputs for the temperature monitoring network. Top left : The predicted map of temperatures using all EOFs components, at the fixed time indicated by the vertical dashed line in the temporal plot below. Top right : The predicted map using only the first 24 components at the same time. Center : The true time series (in black) at a testing station marked by a cross in the maps above, the predicted time series with all EOFs components (in orange) and the predicted time series with the first 24 EOFs components (in green). For visualization purposes, only the first 42 days of the time series are shown. Bottom left: Accuracy plot at the testing station for the model with all EOFs components. Bottom right: Accuracy plot at the testing station for the model with the first 24 EOFs components.

As in the case of the simulated dataset, we performed a comparison between the multiple output strategy, where the spatial coefficients are modelled jointly, and the use of separate regression models to predict each coefficient map (Table 1). Once more, the latter approach results in a higher error, both for the models with all the components and for the model with the first 24 ones. In both cases, the use of a single output strategy results in comparable error rates between the RF and the neural network models. Again, we can conclude that the use of a single network to model jointly the spatial coefficients and the spatio-temporal fields yields better performances, since the algorithm is trained to minimize a loss computed on the final output after the recomposition of the signal.

A variography study was performed on the test data. Figure 8 shows the spatio-temporal semivariograms for the raw test data and for the modelled data and residuals resulting from the models, with all the components and with only the first 24 ones. While both models are able to coherently reconstruct the variability of the raw data, the semivariogram for the model including all the components has a sill, i.e. the value attained by the semivariogram when the model first flattens out, comparable to the one of the raw data. The sill of the semivariogram computed on the modelled data using 24 components is slightly lower, indicating that a certain amount of the variability of the data has not been captured by the model. This is somehow related to the fact that about 5% of the variability of the training data was not explained by the first 24 components, as indicated in Fig. 6. The two semivariograms for the residuals show that the models were able to retrieve most of the spatio-temporal structure of the data. Nonetheless, it can be seen how residuals still show a small temporal correlation, corresponding to the daily cycle of temperature. This is likely because the spatial modeling of each component induces an error, which in the reconstructed field becomes proportional to its corresponding temporal basis. This suggests that even if the spatial component models are correctly modeled, some temporal dependencies may subsist.

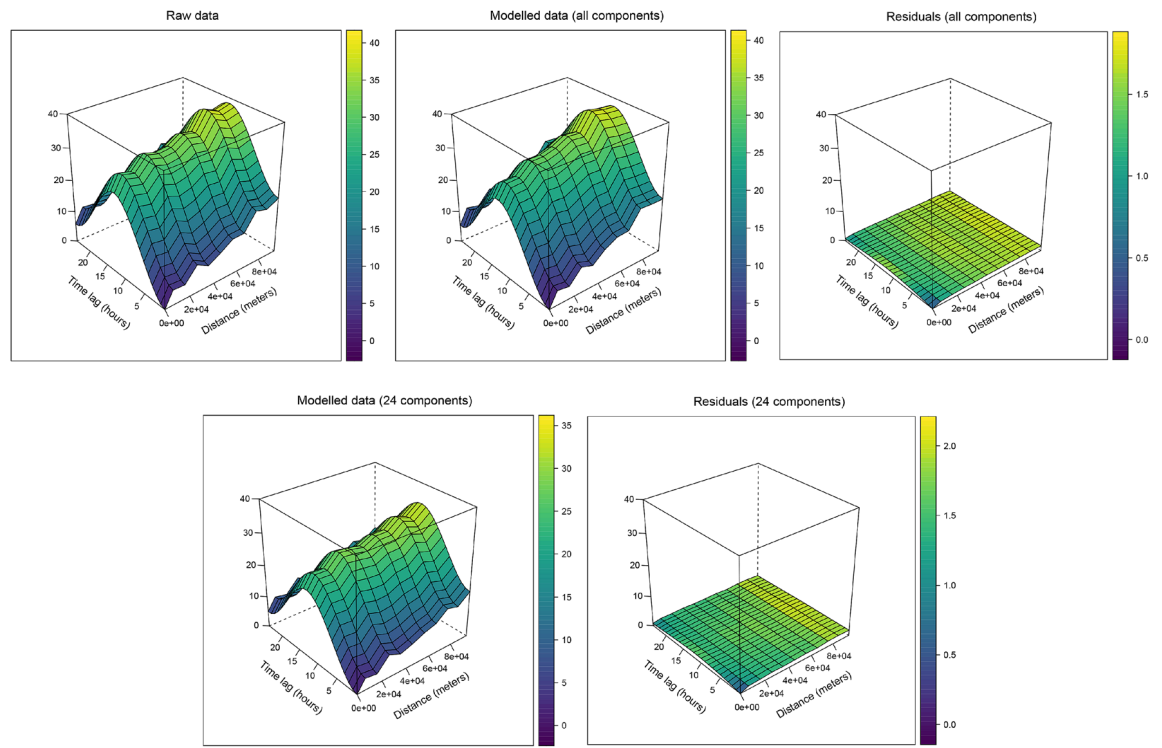

**Figure 8.** Variography for the temperature monitoring network models. Spatio-temporal semivariograms have been computed on the test points. Top row: semivariograms for the raw data (left), the model implemented using all the EOFs components (center) and its residuals (right). Bottom row: semivariograms for the model implemented using the first 24 EOFs components (left) and its residuals (right).

## Discussion

In this paper we introduced a framework for spatio-temporal prediction of climate and environmental data using DL. The proposed approach has two key advantages. First, the decomposition of the spatio-temporal signal into fixed temporal bases and stochastic spatial coefficients allows to fully reconstruct spatio-temporal fields starting from spatially irregularly distributed measurements. Second, while the spatial prediction of the stochastic coefficients can be performed using any regression algorithm, DL algorithms are particularly well suited to solve this problem thanks to their automatic feature representation learning. Furthermore, our framework is able to capture non-linear patterns in the data, as it models spatio-temporal fields as a combination of products of temporal bases by spatial coefficients maps, where the latter are obtained using a non-linear model. Moreover, it can be shown that the basis-function random effects representation induces a valid marginal covariance function without requiring a complete prior knowledge of the phenomena by the modeller, which would instead be necessary in the applications based on traditional geostatistics. Finally, even if the traditional ML and geostatistical techniques could be used to model separately each single spatial coefficient map, the use of a single DL model allows the development of a network structure with multiple outputs to model them all coherently. Besides, the recomposition of the full spatio-temporal field can be executed through an additional layer embedded in the network, allowing to train the entire model to minimize a loss computed directly on the output signal. We showed that the proposed framework succeeds at recovering spatial, temporal and spatio-temporal dependencies in both simulated and real-world data. Furthermore, the proposed framework can eventually be generalized to study other climate fields and environmental spatio-temporal phenomena—e.g. air pollution or wind speed—or to solve missing data imputation problems in spatio-temporal datasets collected by satellites for earth observation or resulting from climate models.

Researchers in the field of ML are becoming more aware of the relevance of their work in tackling climate change issues, environmental risks (e.g., pollution, natural hazards), renewable energy resources assessment, thus contributing to the new emerging field of environmental, and in particular climate, informatics. With this paper, we sought to broaden the agenda of spatio-temporal data analysis and modelling through deep learning.

Being adaptable to every machine learning models, the approach discussed in this paper may enable users interested in measuring the uncertainties of their model output to use methods allowing its explicit estimation, like for example Gaussian Processes. Nonetheless, further research has to be developed to define procedures to quantify the propagation of uncertainty through the diverse steps of the proposed framework. In addition, further studies must be conducted to analyse the consistency of the spatio-temporal predictions provided by this framework and more generally by any other data-driven method with the patterns observed in the real physical models determining the studied phenomena. With this concern, a promising research direction comes from the possibility of integrating physical and data-driven models. These integrated or hybrid models should ideally be constrained by physical laws while ensuring flexibility and adaptation capacity. Finally, additional fundamental studies will be conducted to extend our approach for spatio-temporal forecasting and multivariate analysis.

### Acknowledgements
The research presented in this paper was partly supported by the National Research Program 75 "Big Data" (PNR75, Project No. 167285 "HyEnergy") of the Swiss National Science Foundation (SNSF). Authors are also grateful to Mohamed Laib for the profitable discussions.

### Author contributions
F.A. and F.G. conceived the main conceptual ideas and conduct the residuals analysis. F.A., F.G. and S.R. pre-processed the data. F.A. performed the calculations, visualized the results and wrote the original draft. M.K. carried out the project administration and funding acquisition. All authors discussed the results, provided critical feedback, commented, reviewed and edited the original manuscript, and gave final approval for publication.

### Competing interests
The authors declare no competing interests.

### Additional information
**Correspondence** and requests for materials should be addressed to F.A.

**Reprints and permissions information** is available at www.nature.com/reprints.

**Publisher's note** Springer Nature remains neutral with regard to jurisdictional claims in published maps and institutional affiliations.